\pdfoutput=1
\documentclass{article}

\PassOptionsToPackage{numbers}{natbib}
\usepackage[main,final]{neurips_2026}
\usepackage{microtype}
\usepackage{graphicx}
\usepackage{subcaption}
\usepackage{booktabs}
\usepackage{multirow}
\usepackage[table]{xcolor}
\usepackage{xspace}
\usepackage{caption}  %
\usepackage{algorithm}
\usepackage{algorithmic}
\usepackage{float}    %
\usepackage{hyperref}

\usepackage{amsmath}
\usepackage{amssymb}
\usepackage{mathtools}
\usepackage{amsthm}

\usepackage[capitalize,noabbrev]{cleveref}

\theoremstyle{plain}

\theoremstyle{definition}

\theoremstyle{remark}

\newcommand{\vQ}{\mathbf{Q}}
\newcommand{\vK}{\mathbf{K}}
\newcommand{\vV}{\mathbf{V}}

\newcommand{\vS}{\mathbf{S}}

\newcommand{\vP}{\mathbf{P}}

\newcommand{\vO}{\mathbf{O}}

\newcommand{\vM}{\mathbf{M}}

\definecolor{deepgreen}{rgb}{0.0, 0.5, 0.0}
\definecolor{deepred}{rgb}{0.6, 0.0, 0.0}
\definecolor{darkgreen}{rgb}{0.15, 0.75, 0.15}
\definecolor{cvprblue}{rgb}{0.21,0.49,0.74}
\definecolor{lightblue}{rgb}{0.90, 0.95, 0.99}

\usepackage[disable,textsize=tiny]{todonotes}

\title{SpargeAttention2: Trainable Sparse Attention via Hybrid Top-k+Top-p Masking and Distillation Fine-Tuning}

\author{%
  \textbf{Jintao Zhang}$^{1}$\thanks{Equal contribution.}\quad \textbf{Kai Jiang}$^{1*}$\quad \textbf{Chendong Xiang}$^{1*}$\quad \textbf{Weiqi Feng}$^{1}$\quad \textbf{Yuezhou Hu}$^{2}$\\
  \textbf{Haocheng Xi}$^{2}$\quad \textbf{Shuo Yang}$^{2}$\quad \textbf{Mengfei Xia}$^{3}$\quad \textbf{Jianfei Chen}$^{1}$\quad \textbf{Jun Zhu}$^{1}$\thanks{Corresponding author: \texttt{dcszj@tsinghua.edu.cn}.}\\
  $^{1}$Tsinghua University\quad $^{2}$UC Berkeley\quad $^{3}$Ant Group
}

\hypersetup{hidelinks, pdftitle={SpargeAttention2: Trainable Sparse Attention via Hybrid Top-k+Top-p Masking and Distillation Fine-Tuning}, pdfauthor={Jintao Zhang, Kai Jiang, Chendong Xiang, Weiqi Feng, Yuezhou Hu, Haocheng Xi, Shuo Yang, Mengfei Xia, Jianfei Chen, Jun Zhu}}

\begin{document}

\maketitle

\begin{abstract}
Many training-free sparse attention methods are effective for accelerating diffusion models. Recently, several works suggest that making sparse attention trainable can further increase sparsity while preserving generation quality.
We study three key questions: (1) when do the two common masking rules, i.e., Top-k and Top-p, fail, and how can we avoid these failures? (2) why can trainable sparse attention reach higher sparsity than training-free methods? (3) what are the limitations of fine-tuning sparse attention using the diffusion loss, and how can we address them?
Based on this analysis, we propose SpargeAttention2, a trainable sparse attention method that achieves high sparsity without degrading generation quality. SpargeAttention2 includes (i) a hybrid masking rule that combines Top-k and Top-p for more robust masking at high sparsity, (ii) an efficient trainable sparse attention implementation, and (iii) a distillation-inspired fine-tuning objective to better preserve generation quality during fine-tuning using sparse attention. Experiments on video diffusion models show that SpargeAttention2 reaches $95\%$ attention sparsity and a $16.2\times$ attention speedup while maintaining generation quality, consistently outperforming prior sparse attention methods.
\end{abstract}

\section{Introduction}

\textbf{Motivation and core problem.} Attention efficiency in video diffusion models~\cite{blattmann2023stable,yang2024cogvideox,zheng2024open,kong2024hunyuanvideo,wan2025} is critical because of their long sequence length and $\mathcal O(N^2)$ time complexity of the attention operator. Sparse attention has been shown to work well in diffusion models. For example, SpargeAttention~\cite{zhangspargeattention}, SVG~\cite{xi2025sparse}, and other training-free sparse attention methods~\cite{li2025radial,chen2025sparse} can save a certain portion of attention computation for video generation. More recently, studies show that trainable sparse attention~\cite{zhang2025sla,zhang2025vsa,wu2025vmoba,zhan2025bidirectional} can achieve even higher sparsity after pre-training or fine-tuning.
The core points of sparse attention methods are (i) designing a reasonable sparse masker, i.e., selecting which tokens in the query, key, and value participate in the computation. For trainable sparse attention, it further requires (ii) an efficient, trainable sparse-attention kernel implementation, and (iii) a suitable training objective that enables the trained sparse attention to maintain high generation quality under high sparsity. In this work, we mainly focus on these three aspects.

\begin{figure}[t]
    \centering
    \includegraphics[width=\textwidth]{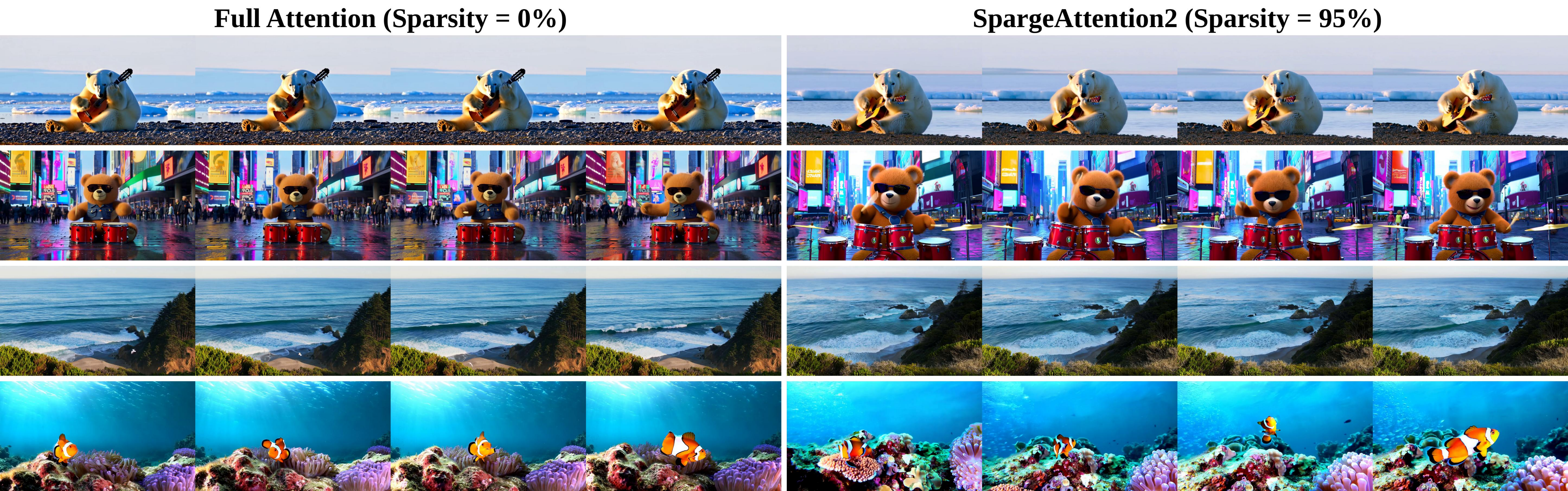}
    \caption{
    Qualitative examples of text-to-video generation.
    We compare the original full-attention model with SpargeAttention2 under high attention sparsity.
    SpargeAttention2 preserves visual quality, temporal coherence, and text--video alignment comparable to full attention,
    while substantially reducing attention computation.
    The prompts used for generation are in Appendix~\ref{app:prompts}
    }
    \label{fig:qualitative_fa_vs_sparge2}
\end{figure}

\textbf{Limitation.} Current trainable sparse attention methods have two main limitations.
\textbf{(L1)} Under very high attention sparsity (e.g., $>90\%$), we observe that both Top-k and Top-p maskers can fail to preserve the most important attention computation. This is closely related to the distribution of each row of the attention weights matrix ($P$), which is often either (i) relatively uniform or (ii) highly skewed.
With a Top-k masker, if the row is close to uniform, the probability is spread over many tokens. Then, keeping a fixed $K$ tokens captures only a small fraction of the total probability, which may miss useful context.
With a Top-p masker, a highly skewed row may satisfy the cumulative-probability threshold with only a few tokens; these tokens can be dominated by attention sinks~\cite{xiao2023efficient,gu2024attention}, causing other informative tokens to be dropped.
\textbf{(L2)} Most existing sparse attention methods fine-tune video diffusion models using prompt--video pairs collected from real-world sources and optimize the standard diffusion loss.
However, in practice, this setting is problematic for widely used open-source video diffusion models, whose pre-training datasets are typically not publicly available (e.g., Wan2.1~\cite{wan2025}).
As a result, it is difficult for the community to collect fine-tuning data that matches the distribution of the original pre-training data.
In this setting, even fine-tuning with \emph{full attention} can noticeably degrade performance relative to the original model. This is because the diffusion loss is data-driven and forces the model to fit the fine-tuning dataset, which is typically lower quality than the original training data.

\textbf{Our approach.}
We propose SpargeAttention2, an accurate and efficient trainable sparse attention method for diffusion models.
To address \textbf{(L1)}, we analyze how Top-k and Top-p masking affect the \emph{information} preserved by sparse attention, especially at very high sparsity. Based on this analysis, we propose a simple and effective unified masker that combines Top-k and Top-p, and works well for both uniform and skewed attention weight distributions.
To address \textbf{(L2)}, inspired by distillation, we introduce a velocity-level distillation loss that aligns the model using sparse attention with a frozen full-attention model during fine-tuning.
Specifically, the velocity distillation loss uses the output of the full-attention model as the supervision signal, which helps maintain the original generation quality even when the fine-tuning data distribution differs from the pre-training distribution.
This design matches the goal of trainable sparse attention, which aims to preserve generation quality while pushing sparsity as high as possible. In contrast, conventional fine-tuning typically aims to enhance or specialize model capabilities.

\textbf{Result.} SpargeAttention2 achieves 95\% attention sparsity, 16.2 $\times$ attention runtime speedup, and up to 4.7 $\times$ end-to-end video generation speedup while maintaining the end-to-end generation quality comparable to full attention, as shown in Figure~\ref{fig:qualitative_fa_vs_sparge2}.

\textbf{Contribution.} Our contributions are summarized as:

(1) We study three key questions in sparse attention for diffusion models: when Top-k and Top-p masking fail, why trainable methods can reach higher sparsity, and why fine-tuning with diffusion loss can be suboptimal. This analysis yields several important insights.

(2) We propose an efficient trainable sparse-attention, SpargeAttention2. It contains (1) a hybrid Top-k and Top-p masker for accurate sparse masking and (2) a distillation-style fine-tuning for trainable sparse attention for enhancing end-to-end generation quality.

(3) SpargeAttention2 achieves 95\% attention sparsity, a 16.2$\times$ attention speedup, and a 4.7$\times$ end-to-end generation speedup without degrading video generation quality, outperforming prior methods.

\section{Preliminaries}

\subsection{Block Sparse Attention}
Let $Q,K,V \in \mathbb{R}^{N\times d}$ be the query, key, and value matrices, where $N$ is the number of tokens and $d$ is the head dimension. Standard attention forms the score matrix $S$ and applies a row-wise softmax to obtain attention weights $P=\mathrm{Softmax}(S)\in\mathbb{R}^{N\times N}$, and produces the attention output $O$.
\[
S={QK^\top}/{\sqrt{d}} \in \mathbb{R}^{N\times N},~~ O = PV \in \mathbb{R}^{N\times d}.
\]
The two matrix multiplications cost $\mathcal{O}(N^2 d)$, which is expensive for large $N$.

Sparse attention reduces this cost by masking out low-importance attention weights. It introduces a binary mask
$M \in \{0,1\}^{N\times N}$ and keep only the selected weights via $P \leftarrow P \odot M$, where $\odot$ denotes element-wise multiplication. A typical choice is thresholding:
$M_{ij}=1$ if $P_{ij}>\tau$ and $M_{ij}=0$ otherwise. When $M_{ij}=0$, we can skip computing the corresponding score and contribution, i.e., the dot product $Q_iK_j^\top$ and the value update $P_{ij}V_j$, where $Q_i\in\mathbb{R}^{d}$ is the $i$-th row of $Q$ and $K_j,V_j\in\mathbb{R}^{d}$ are the $j$-th rows of $K$ and $V$.
In practice, however, fine-grained (element-wise) sparsity maps poorly to modern GPUs. Efficient kernels such as FlashAttention~\citep{dao2023flashattention} therefore exploit \emph{block} structure. Concretely, we partition tensors into tiles:
\begin{align*}
Q=\{\vQ_i\},~~ K=\{\vK_j\},~~ V=\{\vV_j\}, ~~S=\{\vS_{ij}\},~~ P=\{\vP_{ij}\},~~ M=\{\vM_{ij}\}.
\end{align*}
where
$\vQ_i \in \mathbb{R}^{b_q\times d},~
 \vK_j,\vV_j \in \mathbb{R}^{b_{kv}\times d},~ 
\vS_{ij},\vP_{ij},\vM_{ij} \in \mathbb{R}^{b_q\times b_{kv}}$.
Block-sparse attention restricts the mask to be constant within each tile: every $\vM_{ij}$ is either an all-one block (keep) or an all-zero block (drop). 
\[
\vM_{ij}[:,:]=\mathbf{0}\ \Rightarrow\ \text{skip } \vQ_i\vK_j^\top \text{ and } \vP_{ij}\vV_j.
\]
This block-wise gating aligns sparsity with GPU-friendly tiling, enabling practical speedups.

\subsection{Masking for Sparse Attention in Diffusion Models}
Diffusion models do not use autoregressive decoding, so sparse attention is usually implemented in a block-sparse form. The masking problem is therefore to decide, for each block pair $(i,j)$, $\vM_{ij}[:,:]\in\{\mathbf{0},\mathbf{1}\}$.

In practice, forming the full attention weights $P\in\mathbb{R}^{N\times N}$ is prohibitively expensive. To obtain a block mask efficiently, a common approach is to compute a \emph{block-pooled} attention map at the block granularity. Specifically, queries and keys are pooled within each block (e.g., mean pooling over $b_q$ query tokens and $b_{kv}$ key tokens) to produce $\bar{Q}$ and $\bar{K}$. The pooled attention scores and weights can be obtained by:
\[
\bar{S}={\bar{Q}\bar{K}^\top}/{\sqrt{d}},\qquad
\bar{P}=\mathrm{Softmax}(\bar{S}) \in \mathbb{R}^{N/b_q \times N/b_{kv}},
\]
where $\bar{P}_{ij}$ measures the importance of keeping tile $(i,j)$. We define block sparse mask $\bar{M}_{ij}$ as:
\[
\vM_{ij}[:,:]=\mathbf{1} ~\mathrm{or} ~\mathbf{0}\ \Longleftrightarrow\ \bar{M}_{ij}=1 ~\mathrm{or}~ 0.
\]
The block mask is determined by applying Top-k or Top-p to each row of $\bar{P}$:

\textbf{Top-k.}
For each row $i$, keep the $k$\% largest positions in $\bar{P}_{i,:}$:
\[
\bar{M}_{ij}=1 \ \text{if } j \in \rm{Top}\text{-}\rm{k}(\bar{P}_{i,:},k\%),\qquad \bar{M}_{ij}=0 \ \text{otherwise}.
\]
\textbf{Top-p.}
For each row $i$, keep the smallest set of positions whose cumulative probabilities reach $p$\%:
\[
\bar{M}_{ij}=1 \ \text{if } j \in \rm{Top}\text{-}\rm{p}(\bar{P}_{i,:},p\%),\qquad \bar{M}_{ij}=0 \ \text{otherwise},
\]
where $\mathrm{TopP}(\bar{P}_{i,:},p\%)$ denotes the minimal prefix of indices after sorting $\bar{P}_{i,:}$ in descending order such that the summed probability is at least $p$\%.

\subsection{Diffusion Loss}
We adopt the \emph{flow matching}~\cite{lipman2022flow,liu2022flow} formulation as the training objective for diffusion models~\cite{sohl2015deep,ho2020denoising,song2019generative,song2020score}, following the pre-training setup of Wan video models~\cite{wan2025}. Flow matching provides a continuous-time perspective for diffusion modeling, where the generative process is defined by a velocity field rather than discrete denoising steps.
Given a clean image or video latent $x_1$, a noise sample $x_0 \sim \mathcal N(0, I)$, and a time step $t \in [0,1]$ sampled from a predefined schedule, an intermediate latent $x_t$ is constructed as a linear interpolation between $x_0$ and $x_1$:
\begin{equation}
x_t = t x_1 + (1 - t) x_0.
\label{eq:xt}
\end{equation}
The ground-truth velocity is defined as
\begin{equation}
v_t = \frac{d x_t}{d t} = x_1 - x_0.
\end{equation}

The diffusion model $\theta$ is trained to predict this velocity $v_t$ conditioned on the noisy latent $x_t$, timestep $t$, and text prompt $c_{\text{txt}}$. 
Formally, the training objective is formulated as the mean squared error (MSE):
\begin{equation}
\mathcal \rm{Loss} = 
\mathbb E_{x_0, x_1, c_{\text{txt}}, t}
\left[
\left\|
u(x_t, c_{\text{txt}}, t; \theta) - v_t
\right\|^2
\right].
\end{equation}
where $\mathbb E[\cdot]$ denotes expectation taken over the data sample $(x_1, c_{\text{txt}})$, noise sample $x_0$, and timestep $t$.

\section{Analysis} \label{sec:analysis}

\subsection{Error of Sparse Attention}

\paragraph{Notation (one attention row).}
Consider the $i$-th query token. Let $p\in\mathbb{R}^{1\times N}$ denote the attention weights for this row (i.e., the $i$-th row of $P$), let $V\in\mathbb{R}^{N\times d}$ be the value matrix, and let $m\in\{0,1\}^{1\times N}$ be the binary mask for this row (e.g., the $i$-th row of ${M}$). We use $\odot$ for element-wise multiplication.

\paragraph{Sparse-attention error.}
The full-attention output token is $o = pV \in \mathbb{R}^{1\times d}$.
After masking and renormalization, define the retained probability sum
\begin{align}
\tau &= (p\odot m)\mathbf{1}^\top \;=\; \sum_{j=1}^{N} p_j m_j \in \mathbb{R},
\end{align}
and the sparse-attention output $o_s = ({p\odot m}/{\tau}) V \in \mathbb{R}^{1\times d}$.
The error is therefore
\begin{align}
e = o-o_s
&= \left(p - ({p\odot m})/{\tau}\right)V. \label{eq:sparse_error_compact}
\end{align}
The sparse-attention error admits the decomposition
\begin{align}
e
&=\left(
\underbrace{p\odot(1-m)}_{\text{dropped error}}
\;+\;
\underbrace{\left(1-{1}/{\tau}\right)\left(p\odot m\right)}_{\text{renormalization error}}
\right)V,
\label{eq:sparse_error_split}
\end{align}
which separates the dropped contribution (first term) from the renormalization effect (second term).

\subsection{Analysis for Different Cases}

\newcounter{case}
\renewcommand{\thecase}{\arabic{case}}

\newenvironment{casepar}[1]{%
  \refstepcounter{case}%
  \par\vspace{3pt}
  \noindent\textsc{Case \thecase\ (#1).}\ \itshape
}{%
  \par\vspace{3pt}
}

\begin{figure}[h!]
\begin{minipage}[t]{0.48\textwidth}
  \begin{subfigure}[t]{0.47\linewidth}
    \centering
    \fbox{\includegraphics[width=\dimexpr\linewidth-2\fboxsep-2\fboxrule\relax]{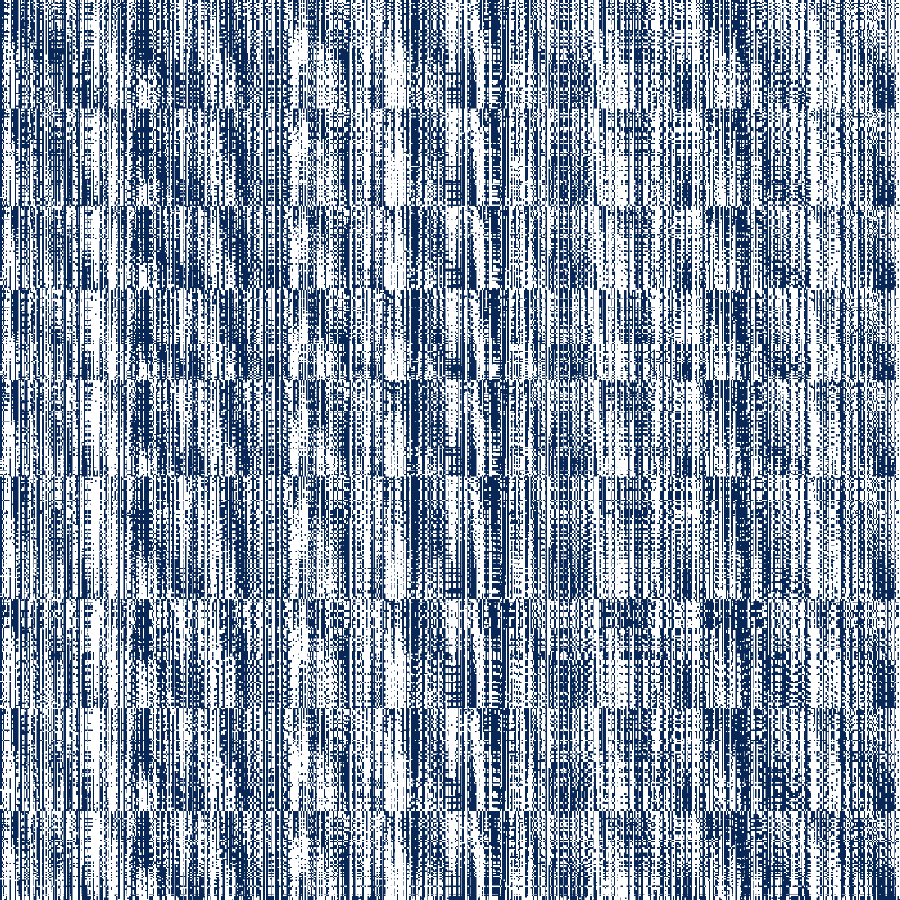}}
    \caption{A uniform $P$. We keep the largest probabilities whose sum reaches $60\%$ in each row.}
    \label{fig:heatmap1a}
  \end{subfigure}\hfill
  \begin{subfigure}[t]{0.47\linewidth}
    \centering
    \fbox{\includegraphics[width=\dimexpr\linewidth-2\fboxsep-2\fboxrule\relax]{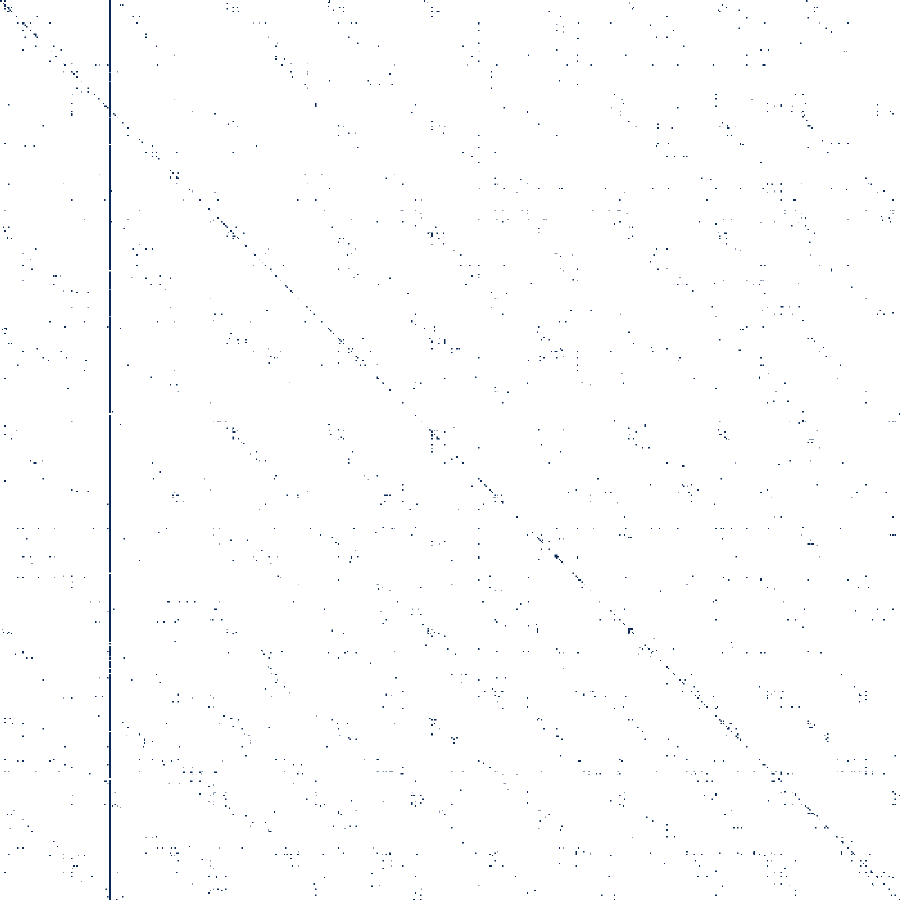}}
    \caption{A skewed $P$, where we keep the largest probabilities whose sum reaches $60\%$ in each row.}
    \label{fig:heatmap1b}
  \end{subfigure}
  \caption{Uniform and skewed heatmap examples for Case~\ref{case1}.}
  \label{fig:heatmaps1}
\end{minipage}%
\hfill%
\setcounter{subfigure}{0}%
\begin{minipage}[t]{0.48\textwidth}
  \begin{subfigure}[t]{0.47\linewidth}
    \centering
    \fbox{\includegraphics[width=\dimexpr\linewidth-2\fboxsep-2\fboxrule\relax]{src/figs/uniformed_heatmap.pdf}}
    \caption{A $P$ before fine-tuning using sparse attention. Each row keeps the largest probabilities whose sum reaches $60\%$.}
    \label{fig:heatmap2a}
  \end{subfigure}\hfill
  \begin{subfigure}[t]{0.47\linewidth}
    \centering
    \fbox{\includegraphics[width=\dimexpr\linewidth-2\fboxsep-2\fboxrule\relax]{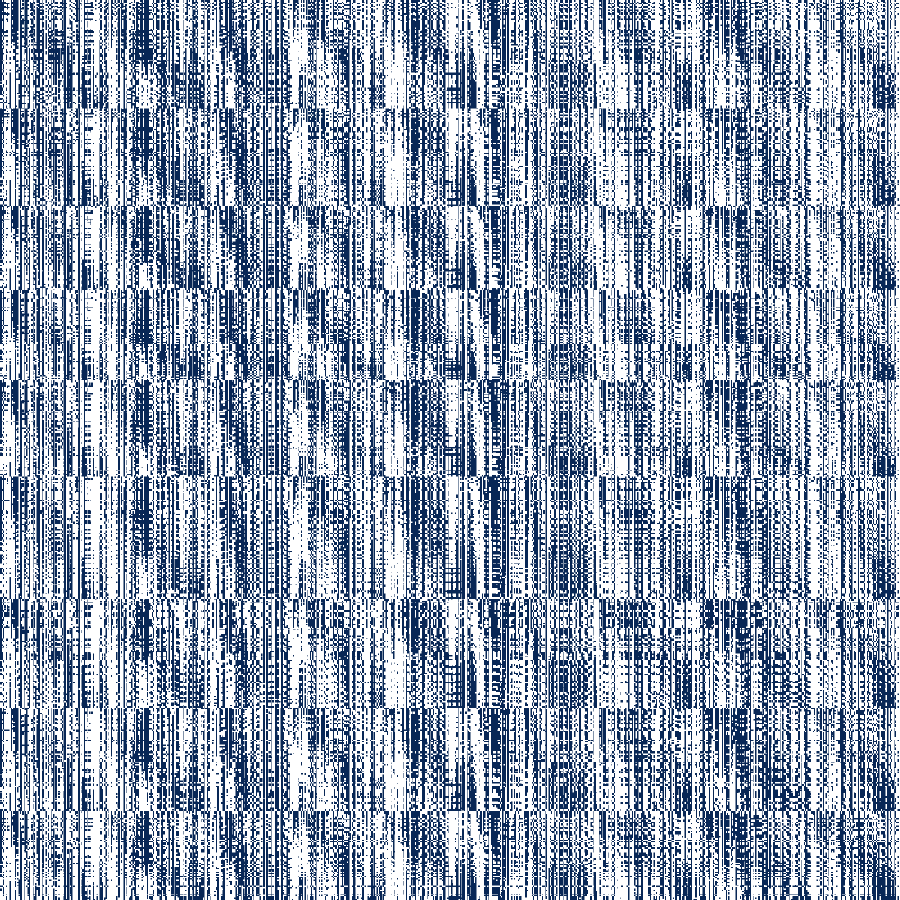}}
    \caption{A sparser $P$ after fine-tuning using sparse attention. Each row keeps the probabilities whose sum reaches $60\%$.}
    \label{fig:heatmap2b}
  \end{subfigure}
  \caption{Heatmaps before and after fine-tuning for Case~\ref{case2}.}
  \label{fig:heatmaps2}
\end{minipage}
\end{figure}

\noindent
\begin{minipage}[t]{0.48\textwidth}
  \centering
  \small
  \setlength{\tabcolsep}{4pt}
  \captionof{table}{L1 error of three masking methods on $P$ with uniform or skewed row distributions.}
  \label{tab:masking_error_analysis}
  \begin{tabular}{lccc}
  \hline
  \textbf{$P$ distribution} & \textbf{Top-k} & \textbf{Top-p} & \textbf{Top-k + Top-p} \\
  \hline
  (a) Uniform  & 0.4150 & \textbf{0.3726} & \textbf{0.3707} \\
  (b) Skewed   & \textbf{0.1664} & 0.2160 & \textbf{0.1671} \\
  \hline
  \end{tabular}
\end{minipage}
\hfill
\begin{minipage}[t]{0.48\textwidth}
  \centering
  \small
  \captionof{table}{Attention sparsity at a 60\% cumulative-probability threshold, and L1 error measured separately at 40\% sparsity, before and after fine-tuning.}
  \label{tab:case2}
  \setlength{\tabcolsep}{8pt}
  \begin{tabular}{lcc}
  \hline
  \textbf{Fine-tuning} & \textbf{Sparsity} & \textbf{L1 Error} \\
  \hline
  (a) Before & 51.3\% & 0.4901 \\
  (b) After  & \textbf{56.9\%} & \textbf{0.4119} \\
  \hline
  \end{tabular}
\end{minipage}

\begin{casepar}{Failure of Top-k and Top-p masking}  \label{case1}

In Figure~\ref{fig:heatmaps1}, we select two representative attention-weight $P$ matrices to analyze the accuracy of different masking strategies. For the left $P$ (Figure~\ref{fig:heatmap1a}), each row has an almost uniform probability distribution. We call it \emph{uniform $P$}. For the right $P$, each row is highly concentrated. We call it \emph{skewed $P$}.
Under the same attention sparsity (i.e., each masking strategy keeps the same number of attention weights), we compare three masking methods: Top-k, Top-p, and their combination (Top-k+Top-p). We measure accuracy by the relative $L1$ distance between the sparse attention output and the full attention output. As shown in Table~\ref{tab:masking_error_analysis}, for uniform $P$, the accuracy satisfies:
\[
\rm{Top}\text{-}\rm{p} ~\approx~ \rm{Top}\text{-}\rm{k}{+}\rm{Top}\text{-}\rm{p} \;>\; \rm{Top}\text{-}\rm{k}.
\]
This is because when the probabilities are spread across many tokens, Top-k keeps only a fixed number of probabilities and may miss many important ones. For example, if a row contains ten probabilities of $0.1$, Top-$20\%$ keeps only two of them, i.e., 2 high-probabilities. This significantly increases the dropped error, i.e., the first term in Equation~\ref{eq:sparse_error_split}.

For skewed $P$, the accuracy satisfies:
\[
\rm{Top}\text{-}\rm{k} ~\approx~ \rm{Top}\text{-}\rm{k}{+}\rm{Top}\text{-}\rm{p} \;>\; \rm{Top}\text{-}\rm{p}.
\]
This is because when the distribution is highly concentrated, Top-p may reach the cumulative threshold with only a few probabilities corresponding to attention sinks~\cite{xiao2023efficient,gu2024attention}. For example, for a row like $[0.6\ \text{(sink)},\,0.2,\,0.1,\,\ldots]$, Top-p$(60\%)$ selects only the sink probabilities and ignores other important probabilities, which increases the error of sparse attention. In contrast, Top-k could select not only the attention sink probabilities.
\end{casepar}

\begin{casepar}{Attention becomes sparser after training} \label{case2}

As shown in Figure~\ref{fig:heatmaps2}, we visualize two heatmaps of $P$ from a diffusion model: (a) before fine-tuning with sparse attention and (b) after fine-tuning with sparse attention. For a fair comparison, we keep the largest probabilities until their sum reaches $60\%$ for each row. Table~\ref{tab:case2} shows that, after sparse-attention fine-tuning, $P$ becomes more sparse (i.e., probabilities are more concentrated). We further compare the attention $L1$ error at the same 40\% sparsity. Table~\ref{tab:case2} shows that the fine-tuned model achieves a smaller error, which helps explain why trainable sparse attention performs better in practice.

This observation also matches the error decomposition in Equation~\ref{eq:sparse_error_split}. Specifically, if fine-tuning makes the attention distribution more concentrated, the \emph{dropped error} and the \emph{renormalization error} will reduce under the same attention sparsity before fine-tuning. The dropped term $(p\odot(1-m))V$ becomes smaller because, when $p$ is more concentrated, the probabilities masked out by $(1-m)$ carry less probability. Meanwhile, the remained probability sum $\tau=(p\odot m)\mathbf{1}^\top$ becomes larger, so the magnitude $\left|1-\frac{1}{\tau}\right|$ decreases, reducing the renormalization term. A simple example could illustrate this effect: Suppose that before fine-tuning,
$p = [0.6,\,0.2,\,0.2],$
and after fine-tuning,
$p = [0.8,\,0.1,\,0.1].$
At $2/3$ sparsity, the sparse attention mask $m=[1,0,0]$. The dropped probability is $p\odot(1-m)$, which equals $[0,0.2,0.2]$ before fine-tuning and $[0,0.1,0.1]$ after fine-tuning. Additionally, the $1/\tau$ will also decrease. As a result, the sparse-attention error is smaller after fine-tuning.
\end{casepar}

\begin{table}[h!]
\centering
\caption{
{Full-attention diffusion fine-tuning degrades alignment under distribution mismatch.}
Without access to the original pre-training data, optimizing the diffusion loss alone leads to consistent degradation in aesthetic quality, vision reward, and VQA accuracy, even when full attention is used.
}
\vspace{4pt}
\label{tab:full_attention_ft}
\small
\setlength{\tabcolsep}{15pt}
\begin{tabular}{l l cccc}
\toprule
Model & Scale-Res. & AQ$\uparrow$ & VR$\uparrow$ & VA$\uparrow$ & VT$\uparrow$ \\
\midrule
Original
& \multirow{2}{*}{1.3B-480p}
& \textbf{64.41} & \textbf{0.1084} & \textbf{81.28} & \textbf{85.80} \\
Fine-tuning
&
& 61.83 & 0.0936 & 75.45 & 80.87 \\
\midrule
Original
& \multirow{2}{*}{14B-720p}
& \textbf{64.66} & \textbf{0.1238} & \textbf{86.15} & \textbf{87.00} \\
Fine-tuning
&
& 63.48 & 0.1181 & 79.86 & 81.95 \\
\bottomrule
\end{tabular}
\end{table}

\begin{casepar}{Diffusion loss failed in fine-tuning} \label{case3}

Table~\ref{tab:full_attention_ft} compares the original pre-trained models with the full-attention same models after fine-tuning using the \emph{standard diffusion-loss-based optimization} adopted by prior sparse-attention methods. Despite keeping full attention, diffusion-loss-based fine-tuning degrades performance across several key metrics for both the 1.3B and 14B models.
This degradation comes mainly from the quality of the fine-tuning data. With diffusion loss, the model is trained to fit the fine-tuning set, so the result strongly depends on the data. Compared with continuous pre-training, the only major change in our fine-tuning is the dataset. If the fine-tuning dataset has similar quality to the pre-training data, the model should keep similar performance after fine-tuning. However, pre-training data are usually closed and high-quality, so it is hard to collect a matching dataset.
In this setting, this degradation is related to the dataset, not to the use of full or sparse attention.
Therefore, fine-tuning with sparse attention will also be affected by this issue. We propose an effective and simple solution in Section~\ref{method}.
\end{casepar}

\section{Method}
\label{method}

\subsection{Hybrid Top-k+Top-p Masking}

To make sparse attention stably work at high sparsity, we need to avoid the two failure conditions in Case~\ref{case1} in Section~\ref{sec:analysis}. In particular, high sparsity attention should not keep a fixed number of tokens for uniform $P$, and should not rely on a fixed cumulative-probability threshold for skewed $P$.
This can be achieved by using Top-k and Top-p masking together. Specifically, for rows of $P$ with a relatively uniform probability distribution, Top-p helps prevent the Top-k failure where a fixed $k$ may keep too few useful tokens. For rows of $p$ with a highly skewed distribution, Top-k helps prevent the Top-p failure where the cumulative threshold can be met by too few tokens corresponding to the attention sink, leading to an ineffective selection. Formally, we can determine the $\bar M = \rm{Top}\text{-}\rm{kp}(\bar P, k\%, p\%)$ as follows.
\begin{align}
  \bar M_{ij} =
\begin{cases}
1, & j \in \mathrm{Top}\text{-}\rm{k}(\bar{P}_{i,:},k\%)\cup \mathrm{Top}\text{-}\rm{p}(\bar{P}_{i,:},p\%),\\
0, & \text{otherwise}.
\end{cases}
\end{align}

\subsection{Velocity Distillation Loss}
Data distribution mismatch introduces additional performance degradation during sparse-attention adaptation. As analyzed in Case~\ref{case3}, even for full-attention models, optimizing the standard diffusion objective under such distribution mismatch can cause significant behavior drift, because the diffusion loss encourages the model to fit the fine-tuning data distribution. This drift directly conflicts with the goal of sparse-attention adaptation, which aims to adapt the new attention structure while keeping the original generation behavior. Therefore, this issue is not caused by sparse attention itself and cannot be resolved by modifying the attention structure alone, but instead requires a different fine-tuning objective.

To address this issue, we replace the data-driven diffusion objective with a \emph{velocity distillation loss} that directly constrains a sparse-attention model to match a frozen full-attention reference model. Instead of using supervision derived from the fine-tuning data, the sparse-attention model is trained to match the diffusion behavior of the original full-attention model.
We adopt a teacher--student setup~\cite{hinton2015distilling}, where the original full-attention diffusion model serves as a frozen teacher, and the sparse-attention model serves as a student.
Both models share the same initialization and differ only in the attention operator.
During training, the teacher and student receive identical inputs: noisy latent $x_t$ constructed following Eq.~\ref{eq:xt}, timestep $t$, and text conditioning $c_{\text{txt}}$.
We then train the student to align its diffusion dynamics with those of the teacher under these identical noisy inputs.
Let $u_{\text{full}}(x_t, c_{\text{txt}}, t)$ and $u_{\text{sparse}}(x_t, c_{\text{txt}}, t)$ denote the teacher's and student's velocity predictions, respectively. We minimize the following velocity distillation loss:
\[
\mathcal L_{\text{VD}} =
\mathbb E_{x_0, x_1, c_{\text{txt}}, t}
\left[
\left\|
u_{\text{sparse}}(x_t, c_{\text{txt}}, t)
-
u_{\text{full}}(x_t, c_{\text{txt}}, t)
\right\|^2
\right].
\]

Under the flow matching framework, the diffusion dynamics are parameterized by the velocity field $u(x_t, c_{\text{txt}}, t)$.
As a result, minimizing the velocity distillation loss directly aligns the sampling dynamics of the teacher and student models.

Overall, velocity distillation uses the teacher's predictions as supervision to guide sparse-attention adaptation. We do not use the standard diffusion loss during fine-tuning; the fine-tuning data are only used to construct noisy inputs $x_t$ for distillation.
This design avoids introducing optimization gradients that push the model toward the mismatched fine-tuning data distribution, thereby significantly reducing behavior drift while enabling stable adaptation to sparse attention under high sparsity.

\subsection{Kernel Implementation And Model Adaptation}

Algorithm~\ref{alg:fwd} shows the kernel implementation of SpargeAttention2.
We denote SpargeAttention2 as an attention operator
$
O = \rm{SpargeAttn2}(Q, K, V, k\%, p\%),
$
which computes sparse-attention outputs using the hybrid Top-k+Top-p masking strategy in Section~\ref{method}.
We implement mask construction and the block-sparse attention forward/backward passes in CUDA, building on FlashAttention. This implementation efficiently skips the masked-out matrix multiplications and softmax computations.

Algorithm~\ref{alg:training} summarizes the procedure for adapting a pre-trained diffusion model to sparse attention using SpargeAttention2. Starting from a diffusion model with full-attention, we replace all attention layers with SpargeAttention2.
The diffusion model using sparse attention is then adapted by minimizing the difference between its velocity predictions and those of a frozen full-attention teacher.

\noindent\resizebox{\textwidth}{!}{%
\begin{minipage}[t]{0.60\textwidth}
\begin{algorithm}[H]
    \caption{SpargeAttention2 Implementation.}
    \label{alg:fwd}
    \begin{algorithmic}[1]
    \STATE {\bf Input:} {Matrices $Q, K, V \in \mathbb{R}^{N \times d}$, $b_q, b_{kv}$, $k$\%, $p$\%}.
    \STATE {Divide $Q$ to $T_m = N / b_q$ blocks $\{\vQ_i\}$} ;
    \STATE {Divide $K, V$ to $T_n = N / b_{kv}$ blocks $\{\vK_i\}$, $\{\vV_i\}$} ;
    \STATE {$\bar P = {\rm softmax}({\rm pool}(Q){\rm pool}(K)^\top / \sqrt{d})$} ;
    \STATE $\bar M_1=\rm{Top}\text{-}\rm{k}(\bar P, k\%)$ , ~~$\bar M_2=\rm{Top}\text{-}\rm{p}(\bar P, p\%)$ ;
    \STATE {$\bar M= \bar M_1 \cup \bar M_2$ } ;
    \FOR {$i=1$ {\bf to} $T_m$}
        \STATE Initialize $m_{i,0}=-\infty$, $l_{i,0}=0$, $\vO_{i,0}=0$.
        \FOR {$j=1$ {\bf to} $T_n$}
            \IF {$\bar M[i,j]=1$}
                \STATE {$\vS_{ij} = \vQ_i \vK_j^\top / \sqrt{d}$ ;
                \STATE $m_{ij} = {\rm max}(m_{i, j-1}, {\rm rowmax}(\vS_{ij}))$} ;
                \STATE {$\vP_{ij}=\exp(\vS_{ij}-m_{ij})$} ;
                \STATE $l_{ij}=e^{m_{i,j-1}-m_{ij}} l_{i,j-1} + {\rm rowsum}(\vP_{ij})$ ;
                \STATE {$\vO_{ij} = {\rm diag}(e^{m_{i,j-1}-m_{ij}}) \vO_{i,j-1} + \vP_{ij} \vV_j$} ;
            \ELSE
                \STATE Carry $m_{ij},l_{ij},\vO_{ij}$ from tile $j-1$.
            \ENDIF
        \ENDFOR
        \STATE $\vO_i={\rm diag}(l_{i,T_n})^{-1}\vO_{i,T_n}$ ;
    \ENDFOR
    \STATE \textbf{return} $O=\{\vO_i\}$ ;
    \end{algorithmic}
\end{algorithm}
\end{minipage}
\hspace{0.02\textwidth}
\begin{minipage}[t]{0.60\textwidth}
\begin{algorithm}[H]
    \caption{Adapting Diffusion Models with SpargeAttention2 via Velocity Distillation.}
    \label{alg:training}
    \begin{algorithmic}[1]
    \STATE {\bf Input:} {Pre-trained diffusion model $\theta_{\rm full}$, sparsity hyperparameters $k\%, p\%$, training data $\mathcal D$.}
    \STATE {\bf Output:} {Sparse-attention model $\theta_{\rm sparse}$.}
    \STATE Initialize $\theta_{\rm sparse} \leftarrow \theta_{\rm full}$ and freeze $\theta_{\rm full}$.
    \STATE Replace all attention layers in $\theta_{\rm sparse}$ with $\rm{SpargeAttn2}(\cdot,\cdot,\cdot,k\%,p\%)$ (Alg.~\ref{alg:fwd}).
    \FOR{each training iteration}
        \STATE Sample $(x_1, c_{\rm txt}) \sim \mathcal D$, noise $x_0 \sim \mathcal N(0,I)$, and select a timestep $t \in [0,1]$ according to a predefined schedule.
        \STATE Construct noisy latent: \;$x_t = t x_1 + (1-t)x_0$.
        \STATE Compute teacher velocity with full attention:
        \STATE \hspace{6pt} $u_{\rm full} = u_{\theta_{\rm full}}(x_t, c_{\rm txt}, t)$.
        \STATE Compute student velocity with SpargeAttention2:
        \STATE \hspace{6pt} $u_{\rm sparse} = u_{\theta_{\rm sparse}}(x_t, c_{\rm txt}, t)$.
        \STATE Velocity distillation loss:
        \STATE \hspace{6pt} $\mathcal L_{\rm VD} = \|u_{\rm sparse} - u_{\rm full}\|^2$.
        \STATE Update $\theta_{\rm sparse}$ by minimizing $\mathcal L_{\rm VD}$.
    \ENDFOR
    \STATE \textbf{return} $\theta_{\rm sparse}$.
    \end{algorithmic}
\end{algorithm}
\end{minipage}
}

\section{Experiments}
\subsection{Setup}

\textbf{Models and dataset.}
We conduct video generation experiments using the Wan2.1~\cite{wan2025} under two configurations:
Wan2.1-1.3B at 480p resolution and Wan2.1-14B at 720p resolution.
For training, we use a private video dataset consisting of 3,000 videos, each approximately 5 seconds long, collected from publicly available sources.
All videos are stored at a native resolution of 720p.
For the 1.3B model, videos are resized to 480p during training, while for the 14B model, both training and evaluation are performed at 720p resolution.
To obtain text--video pairs, we automatically generate captions for each video using Qwen3-VL-Flash~\cite{Qwen3-VL}. For evaluation, we adopt the prompts provided by VBench~\cite{huang2024vbench} as text inputs for video generation.

\textbf{Baselines and ablations.}
We compare our method with representative trainable sparse attention approaches for diffusion models, including VSA~\cite{zhang2025vsa}, VMoBA~\cite{wu2025vmoba}, SLA~\cite{zhang2025sla}, and SpargeAttention~\cite{zhangspargeattention}.
In addition, we conduct controlled ablation studies by modifying only one component at a time.
Specifically, we study three aspects:
(1) \emph{Sparse masker design}, by replacing the unified Top-k+Top-p masker with Top-k-only or Top-p-only variants;
(2) \emph{Effect of training}, by comparing trainable sparse attention with a training-free variant where sparse-attention parameters are frozen;
and (3) \emph{Training objective}, by replacing the proposed velocity distillation loss with standard diffusion-loss-based training.

\textbf{Metrics.}
For video generation quality, we report Imaging Quality (IQ), Overall Consistency (OC), and Aesthetic Quality (AQ) from the VBench benchmark~\cite{huang2024vbench}, together with Vision Reward (VR)~\cite{xu2024visionreward} and VQA accuracy (VA and VT)~\cite{liu2024evalcrafter}, where VA and VT denote VQA-a and VQA-t, respectively, following prior work~\cite{zhang2025sla, wu2025vmoba}. All training hyper-parameters and sparse-attention settings are provided in Appendix~\ref{app:hyperparameters}.
For efficiency evaluation, we report attention latency and end-to-end generation latency in seconds (s), measured on an RTX 5090 GPU.

\subsection{Effectiveness}

\begin{figure}[t]
    \centering
    \includegraphics[width=\textwidth]{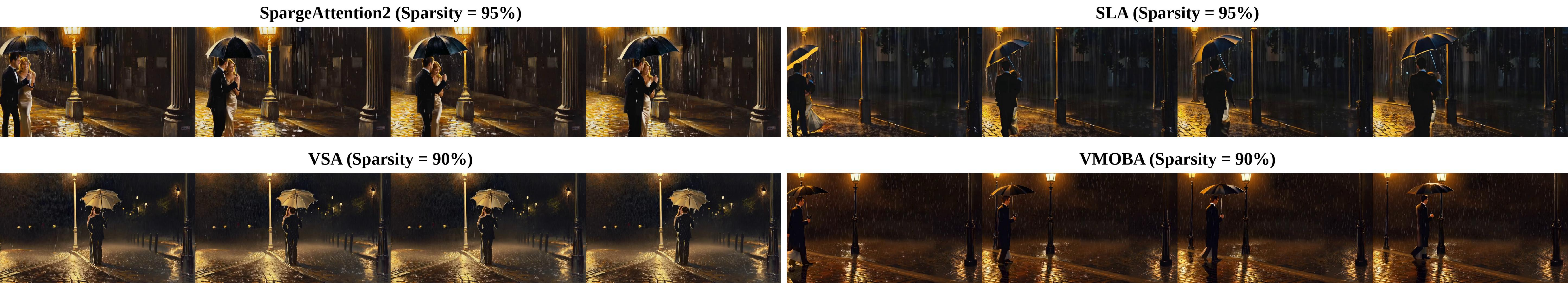}
    \caption{
    A representative example of text-to-video generation under high attention sparsity,
    evaluated on Wan2.1-14B at 720p.
    SpargeAttention2 produces a semantically correct video.
    In contrast, SLA and VSA produce videos in which the male character walks backward,
    while VMoBA fails to generate the female character specified in the prompt.
    The prompt used for generation is in Appendix~\ref{app:prompts}
    }
    \label{fig:qualitative_sparse}
\end{figure}

We evaluate SpargeAttention2 against prior trainable sparse-attention methods on Wan2.1 under a high attention sparsity.
Results on Wan2.1-1.3B at 480p and Wan2.1-14B at 720p are reported in Tables~\ref{tab:effectiveness_1p3b} and~\ref{tab:effectiveness_14b}, respectively.
Complete VBench scores and results on a private MM-DiT model are reported in Appendices~\ref{sec:complete-vbench} and~\ref{sec:mmdit}.
Across both settings, SpargeAttention2 outperforms the sparse-attention baselines on all six reported quality metrics and retains generation quality close to full attention.
In contrast, existing sparse-attention baselines exhibit noticeable degradation under the same or even lower sparsity levels.
These results indicate that SpargeAttention2 performs robustly across different model sizes and resolutions.
Figure~\ref{fig:qualitative_sparse} provides a qualitative comparison.

\begin{table}[t]
\centering
\caption{Effectiveness comparison on Wan2.1-1.3B at 480p resolution. Bold indicates the best result among sparse-attention methods.}
\resizebox{\textwidth}{!}{%
\small
\setlength{\tabcolsep}{5.5pt}
\begin{tabular}{lccccccccc}
\toprule
Method
& IQ$\uparrow$
& OC$\uparrow$
& AQ$\uparrow$
& VR$\uparrow$
& VQA-a$\uparrow$
& VQA-t$\uparrow$
& Sparsity$\uparrow$
& Attn Time$\downarrow$
& E2E Time$\downarrow$ \\
\midrule
Full Attention
& 63.67 & 20.27 & 64.41 & 0.1084 & 81.28 & 85.80 & 0\% & 97s & 159s \\
\midrule
SpargeAttn
& 35.28 & 13.41 & 40.53 & -0.1398 & 3.258 & 0.608 & 89\% & 12.6s & 74.6s \\
VSA
& 59.57 & 19.27 & 50.60 & -0.0881 & 33.35 & 48.36 & 90\% & 25s & 87s \\
VMoBA
& 65.31 & 20.82 & 64.14 & 0.0936 & 78.99 & 86.69 & 90\% & 36s & 98s \\
SLA
& 63.14 & 21.09 & 62.91 & 0.0881 & 72.66 & 80.51 & \textbf{95\%} & 11s & 73s \\
\textbf{SpargeAttn2}
& \textbf{67.68}
& \textbf{21.57}
& \textbf{65.05}
& \textbf{0.1010}
& \textbf{83.86}
& \textbf{87.73}
& \textbf{95\%}
& \textbf{6s} & \textbf{68s} \\
\bottomrule
\end{tabular}%
}
\label{tab:effectiveness_1p3b}
\end{table}

\begin{table}[t]
\centering
\caption{Effectiveness comparison on Wan2.1-14B at 720p resolution. Bold indicates the best result among sparse-attention methods.}
\resizebox{\textwidth}{!}{%
\small
\setlength{\tabcolsep}{5.5pt}
\begin{tabular}{lccccccccc}
\toprule
Method
& IQ$\uparrow$
& OC$\uparrow$
& AQ$\uparrow$
& VR$\uparrow$
& VQA-a$\uparrow$
& VQA-t$\uparrow$
& Sparsity$\uparrow$
& Attn Time$\downarrow$
& E2E Time$\downarrow$ \\
\midrule
Full Attention
& 68.01 & 22.44 & 64.66 & 0.1238 & 86.15 & 87.00 & 0\% & 2550s & 3043s \\
\midrule
SpargeAttn
& 38.46 & 16.26 & 42.16 & -0.1306 & 9.926 & 4.451 & 86\% & 415s & 908s \\
VSA
& 64.03 & 21.27 & 63.37 & 0.1074 & 77.63 & 85.60 & 90\% & 651s & 1144s \\
VMoBA
& 67.18 & 20.85 & 63.64 & 0.1117 & 81.66 & 83.21 & 90\% & 832s & 1325s \\
SLA
& 64.43 & 20.89 & 61.89 & 0.1078 & 76.90 & 82.60 & \textbf{95\%} & 285s & 778s \\
\textbf{SpargeAttn2}
& \textbf{69.08}
& \textbf{21.62}
& \textbf{64.92}
& \textbf{0.1149}
& \textbf{85.21}
& \textbf{87.48}
& \textbf{95\%}
& \textbf{157s} & \textbf{650s} \\
\bottomrule
\end{tabular}%
}
\label{tab:effectiveness_14b}
\end{table}

\subsection{Efficiency}

We evaluate the efficiency of SpargeAttention2 on Wan2.1 under high attention sparsities, focusing on both attention operator latency and end-to-end video generation time.
Results on Wan2.1-1.3B at 480p and Wan2.1-14B at 720p are reported in Tables~\ref{tab:effectiveness_1p3b} and~\ref{tab:effectiveness_14b}, respectively.

At approximately 95\% sparsity, SpargeAttention2 is the only method that simultaneously achieves strong generation quality and substantial efficiency gains.
In contrast, other sparse attention baselines are significantly slower than SpargeAttention2 and exhibit clear degradation in generation quality. Notably, SpargeAttention2 achieves higher video generation quality than all baselines, even at higher attention sparsity.
For efficiency, on Wan2.1-1.3B at 480p, SpargeAttention2 reduces attention latency from 97s to 6s, achieving a 16.2$\times$ speedup over full attention.
It is 1.8$\times$ faster than SLA and more than 4$\times$ faster than VSA and VMoBA, while also delivering clearly superior generation quality.
This reduction in attention cost results in an end-to-end generation speedup from 159s to 68s, corresponding to a 2.3$\times$ overall acceleration.
Similar trends are observed on Wan2.1-14B at 720p.
SpargeAttention2 reduces attention latency from 2550s to 157s, achieving a 16.2$\times$ speedup over full attention.
Compared with prior sparse-attention methods, it is 1.8$\times$ faster than SLA and more than 4$\times$ faster than VSA and VMoBA, while maintaining generation quality comparable to or better than full attention.
End-to-end generation time is reduced from 3043s to 650s, reaching a 4.7$\times$ speedup.

\begin{table}[t]
\centering
\small
\caption{
Ablation studies on SpargeAttention2.
``--VD'' replaces velocity distillation with standard diffusion fine-tuning. The 1.3B and 14B ablations use 500 and 100 training steps, respectively (Appendix~\ref{app:hyperparameters}).
VQA denotes the overall score combining VQA-a and VQA-t.
}
\label{tab:ablation_combined}
\vspace{4pt}
\resizebox{\textwidth}{!}{%
\begin{tabular}{lcccccccccc}
\toprule
& \multicolumn{5}{c}{\textbf{Wan2.1-1.3B-480p}} & \multicolumn{5}{c}{\textbf{Wan2.1-14B-720p}} \\
\cmidrule(lr){2-6} \cmidrule(lr){7-11}
Variant & IQ & OC & AQ & VR & VQA & IQ & OC & AQ & VR & VQA \\
\midrule
\textbf{SpargeAttn2} & \textbf{67.68} & \textbf{21.57} & \textbf{65.05} & \textbf{0.1010} & 86.73 & \textbf{68.41} & 21.06 & \textbf{65.02} & \textbf{0.1119} & \textbf{88.22} \\
Top-k only           & 65.84 & 21.51 & 64.57 & 0.0916 & \textbf{86.90} & 65.24 & 20.64 & 63.99 & 0.0935 & 84.25 \\
Top-p only           & 60.56 & 21.12 & 60.12 & 0.0312 & 62.57 & 63.37 & \textbf{21.33} & 63.62 & 0.1090 & 86.43 \\
Training-free        & 53.18 & 19.87 & 48.93 & -0.0650 & 20.40 & 62.17 & 19.41 & 57.01 & -0.0272 & 45.85 \\
--VD                 & 67.23 & 21.26 & 63.34 & 0.0939 & 85.05 & 66.40 & 20.56 & 64.59 & 0.1082 & 85.05 \\
\bottomrule
\end{tabular}
}
\end{table}

\subsection{Ablation}

We analyze the contribution of individual design choices in SpargeAttention2 through ablation studies, focusing on the sparse masker, trainability, and training objective.
Results are reported in Table~\ref{tab:ablation_combined}.

\textbf{Sparse masker design.}
We compare the proposed hybrid Top-k/Top-p masking with variants that use only Top-k or only Top-p masking.
As shown in Table~\ref{tab:ablation_combined}, the unified Top-k+Top-p masker consistently achieves the best overall generation quality and alignment across both model scales, validating its robustness under high sparsity.

\textbf{Effect of training.}
We evaluate the impact of training sparse attention by comparing SpargeAttention2 with a training-free variant.
Table~\ref{tab:ablation_combined} shows that disabling training leads to substantial degradation in generation quality and alignment for both the 1.3B and 14B models, highlighting the necessity of adapting sparse attention under high sparsity.

\textbf{Training objective.}
We examine the role of the training objective by replacing the proposed velocity distillation loss with standard diffusion loss.
As shown in Table~\ref{tab:ablation_combined}, diffusion-loss-based fine-tuning consistently underperforms velocity distillation.
This confirms the effectiveness of the proposed velocity distillation for sparse-attention adaptation.

\subsection{Additional Evaluations}
Complete VBench evaluation in Appendix~\ref{sec:complete-vbench} confirms that SpargeAttention2 achieves the highest Quality, Semantic, and Total scores among the evaluated sparse-attention methods on both Wan2.1 models. On a private 720p MM-DiT model (Appendix~\ref{sec:mmdit}), SpargeAttention2 outperforms SLA on all three VBench scores, with a Total score of 81.46 versus 81.84 for full attention.

\section{Related Work}

Sparse attention methods can be grouped by whether they require training. First, training-free approaches~\citep{gao2024seerattention,xi2025sparse,zhangspargeattention,ribar2023sparq,yang2025sparse,li2025radial,chen2025sparse,lai2025flexprefill,zhang2023h2o,xiao2023efficient,jiang2024minference,tang2024quest,zhu2025tactic,lin2025twilight,xu2025xattention,xia2025training,chen2025re,zhang2025fast,yang2024post} reduce inference cost by applying a test-time attention mask. Among them, vAttention~\cite{desai2025vattention} uses a hybrid of Top-k and random sampling, which differs from our Top-k and Top-p hybrid and is not designed for diffusion models. Second, trainable sparse attention methods~\citep{zhang2025vsa,wu2025vmoba,zhang2025sla,zhan2025bidirectional,zhou2025trainable,lu2025moba,yuan2025native,liu2025deepseek,cai2025mixture,liu2025fpsattention,sun2025vorta,tan2025dsv,ding2023longnet} enhance attention sparsity by directly using sparse attention during training. 
Some methods in the second category are designed for diffusion models, and SpargeAttention2 belongs to this group. Among them, SpargeAttention2 achieves state-of-the-art performance.

SLA2~\citep{ConcurrentWork1} combines learnable sparse-linear routing with quantization-aware training; SVG-EAR~\citep{zhou2026svg} uses error-aware routing and parameter-free linear compensation.

Low-bit attention~\citep{zhang2025sageattention,zhang2025sageattention2,zhang2025sageattention2++,zhang2025sageattention3,zhang2026sagebwd}, accurate INT8 training~\citep{zhang2025accurate}, and weight-sensitivity analysis~\citep{hu2025identifying} provide complementary efficiency techniques. Quant VideoGen~\citep{xi2026quant} targets KV-cache quantization, while 6Bit-Diffusion~\citep{su2026bit} studies mixed-precision video diffusion. Broader acceleration directions include diffusion distillation~\citep{zheng2025large,zheng2026causal,yang2026turbot2va}, model stitching~\citep{sun2026hybridstitch}, speculative decoding~\citep{hu2026speculative}, and TurboDiffusion~\citep{zhang2025turbodiffusion}. Resolution extrapolation~\citep{zhao2025ultravico,zhao2025ultraimage}, interactive video generation~\citep{zhang2026vidu,zhang2026vidus2}, and streaming serving~\citep{jiang2026turboserve} further motivate efficient attention at scale. A broader taxonomy is given by the efficient-attention survey~\citep{zhangefficient}.

\section{Conclusion}

In this paper, we analyze key challenges in sparse attention for diffusion models and propose SpargeAttention2. It is an efficient and accurate trainable sparse attention method that achieves high sparsity without degrading video generation quality. Specifically, by combining a hybrid Top-k and Top-p sparse masking, an efficient implementation, and a distillation-style fine-tuning method, SpargeAttention2 achieves very high sparsity while preserving generation quality, surpassing baselines. SpargeAttention2 achieves 95\% attention sparsity, 16.2 $\times$ attention runtime speedup, and up to 4.7 $\times$ end-to-end video generation speedup.

\begin{ack}
This work was supported by the National Natural Science Foundation of China (Nos. 62550004, 62595773, and 62376131) and the Ministry of Education of China (No. JYB2025XDXM101).
This work was supported by Ant Group Research Fund.
The authors declare no other competing interests.
\end{ack}

\bibliographystyle{unsrt}
\bibliography{main,camera-ready-references}

\clearpage
\appendix
\section{Additional Experiments}
\label{app:additional-experiments}

\subsection{Complete VBench Evaluation}
\label{sec:complete-vbench}
Table~\ref{tab:complete-vbench} reports the VBench Quality, Semantic, and Total scores on Wan2.1-1.3B and Wan2.1-14B. SpargeAttention2 achieves the highest scores among the evaluated sparse-attention methods on both models.

\begin{table}[htbp]
\centering
\caption{Complete VBench evaluation on Wan2.1. Higher scores are better. Bold indicates the best score among the sparse-attention methods.}
\label{tab:complete-vbench}
\begin{tabular}{lrrrr}
\toprule
Method & Quality & Semantic & Total & Sparsity \\
\midrule
\multicolumn{5}{c}{Wan2.1-1.3B} \\
\midrule
Full Attention & 83.95 & 82.38 & 83.63 & 0\% \\
SpargeAttn & 63.10 & 35.84 & 57.65 & 89\% \\
VSA & 77.21 & 64.32 & 74.63 & 90\% \\
VMoBA & 80.03 & 78.92 & 79.81 & 90\% \\
SLA & 81.24 & 78.79 & 80.75 & 95\% \\
SpargeAttention2 & \textbf{83.86} & \textbf{79.25} & \textbf{82.94} & 95\% \\
\midrule
\multicolumn{5}{c}{Wan2.1-14B} \\
\midrule
Full Attention & 83.27 & 80.20 & 82.66 & 0\% \\
SpargeAttn & 71.93 & 35.34 & 64.61 & 86\% \\
VSA & 80.63 & 76.54 & 79.81 & 90\% \\
VMoBA & 81.27 & 79.57 & 80.93 & 90\% \\
SLA & 79.97 & 76.33 & 79.24 & 95\% \\
SpargeAttention2 & \textbf{83.09} & \textbf{79.71} & \textbf{82.42} & 95\% \\
\bottomrule
\end{tabular}
\end{table}

\subsection{Evaluation on A Private MM-DiT Model}
\label{sec:mmdit}
We additionally evaluate SpargeAttention2 on a private MM-DiT model with joint multimodal attention for 720p video generation. We use the original full-attention model as the frozen teacher and follow the Wan2.1-14B fine-tuning setting. The hybrid masker uses $k=0.03$ and $p=0.15$, targeting 95\% sparsity. Table~\ref{tab:mmdit-vbench} compares SpargeAttention2 with full attention and SLA. SpargeAttention2 outperforms SLA on all three VBench scores and obtains a Total score of 81.46, compared with 81.84 for full attention.

\begin{table}[htbp]
\centering
\caption{VBench evaluation on a private 720p MM-DiT model. Higher scores are better. Bold indicates the better score between the two sparse-attention methods.}
\label{tab:mmdit-vbench}
\begin{tabular}{lrrrr}
\toprule
Method & Quality & Semantic & Total & Sparsity \\
\midrule
Full Attention & 81.55 & 83.01 & 81.84 & 0\% \\
SLA & 80.02 & 73.90 & 78.80 & 95\% \\
SpargeAttention2 & \textbf{81.43} & \textbf{81.56} & \textbf{81.46} & 95\% \\
\bottomrule
\end{tabular}
\end{table}

\section{Hyper-Parameters}
\label{app:hyperparameters}
Unless otherwise stated, all models are trained for 500 steps.
We use a batch size of 64 for Wan2.1-1.3B at 480p resolution and 16 for Wan2.1-14B at 720p resolution.
To reduce computational cost, ablation studies on the 14B model are conducted for 100 training steps, while all main results are reported using models trained for 500 steps.

\textbf{Sparse Attention Settings.}
We calibrate Top-k and Top-p to achieve a target sparsity of approximately 95\% across model scales.
Specifically, for Wan2.1-1.3B at 480p resolution, we use Top-k = 0.03 and Top-p = 0.2.
while for Wan2.1-14B at 720p resolution, we use Top-k = 0.03 and Top-p = 0.16. We set $b_q=128$ and $b_{kv}=64$, respectively.

\textbf{Ablation Settings.}
For ablation studies on sparse masker design, we vary Top-k or Top-p individually while keeping all other settings unchanged.
In the Top-k ablation, we set Top-k = 0.05 for both Wan2.1-1.3B and Wan2.1-14B, resulting in approximately 95\% sparsity.
In the Top-p ablation, we use Top-p = 0.4 for Wan2.1-1.3B at 480p and Top-p = 0.3 for Wan2.1-14B at 720p, corresponding to sparsity levels of 94\% and 93\%, respectively.

\section{Prompts for Qualitative Visualizations}
\label{app:prompts}
This subsection lists the text prompts used to generate the qualitative video samples shown in the main paper.

\paragraph{Prompts for Figure~\ref{fig:qualitative_fa_vs_sparge2}.\\} 

From top to bottom, the prompts are:

\begin{itemize}
    \item 
    \textit{A large polar bear sitting on a rocky Arctic shoreline, casually playing an acoustic guitar with its massive paws. 
    The bear has thick, white fur glistening under soft golden sunlight, a curious expression, and gentle eyes focused on the instrument. 
    Melodic notes seem to ring out across the tundra. 
    Behind, ice floes drift in calm blue water beneath a clear pastel sky. 
    Natural daylight, medium shot from the front, slight low angle emphasizing the bear’s size. 
    Slow camera pan from left to right. 
    Realistic wildlife style with a whimsical twist.}

    \item 
    \textit{A fluffy brown teddy bear with a cheerful expression is energetically playing a red drum kit in the heart of New York City’s Times Square. Bright neon billboards and towering digital screens flash colorful advertisements all around, casting vibrant reflections on the wet pavement. The teddy bear, wearing tiny sunglasses and a denim vest, rhythmically bangs the drums with animated motion, cymbals shimmering. Pedestrians stop to watch in delight, capturing the whimsical scene on their phones. Dynamic camera circling the bear, wide-angle view emphasizing the bustling urban spectacle and lively performance. Cartoon-style 3D animation, high detail, vivid colors.}

    \item 
    \textit{Pacific Coast, Carmel-by-the-Sea, scenic ocean shoreline with rolling waves gently crashing against rocky outcrops and sandy coves. 
    The deep blue Pacific Ocean stretches to the horizon under a soft golden-hour sky, with seagulls gliding above the surf. 
    Waves foam and ripple over smooth stones and tide pools teeming with marine life. 
    Coastal pine trees and native shrubs line the bluffs, framing the pristine beach. 
    Gentle wave motion, sparkling water surface, and a calm breeze enhance the serene atmosphere. 
    Wide-angle landscape shot, natural daylight, realistic detail, peaceful coastal ambiance.}
    
    \item 
    \textit{A vibrant clownfish with bright orange body and white stripes edged in black darts playfully through a lush coral reef. 
    It weaves gracefully between swaying sea anemones and colorful coral formations in vivid pinks, purples, and blues. 
    Sunlight filters through the crystal-clear turquoise water, creating shimmering patterns on the ocean floor. 
    Small bubbles rise as the fish flicks its tail, exploring its intricate underwater home. 
    Wide-angle underwater shot with natural lighting, showcasing rich detail and lively marine biodiversity. 
    Gentle current moves the corals softly.}

\end{itemize}

\paragraph{Prompt for Figure~\ref{fig:qualitative_sparse}.}

\begin{itemize}
    \item 
    \textit{An oil painting of a couple in elegant formal evening wear walking home as a sudden heavy downpour surrounds them, 
    clutching umbrellas that barely shield them from the rain. 
    The man wears a sleek black tuxedo, the woman a flowing satin gown shimmering with raindrops, 
    her hair slightly damp and clinging to her shoulders. 
    Rain streaks through the air in silvery lines under dim streetlights, puddles rippling at their feet. 
    Warm glows from nearby lampposts reflect on wet cobblestones, creating a romantic, cinematic atmosphere. 
    Loose brushstrokes and rich textures evoke emotion and movement. 
    Medium shot, side view, captured from a slight distance.}
\end{itemize}

\clearpage
\section*{NeurIPS Paper Checklist}

\begin{enumerate}

\item {\bf Claims}
    \item[] Question: Do the main claims made in the abstract and introduction accurately reflect the paper's contributions and scope?
    \item[] Answer: \answerYes{}
    \item[] Justification: Both the Abstract and the Introduction indicate the scope and contributions of this paper.

\item {\bf Limitations}
    \item[] Question: Does the paper discuss the limitations of the work performed by the authors?
    \item[] Answer: \answerNo{}
    \item[] Justification: The paper evaluates video diffusion models but does not include a dedicated discussion of the limitations of the proposed method.

\item {\bf Theory assumptions and proofs}
    \item[] Question: For each theoretical result, does the paper provide the full set of assumptions and a complete (and correct) proof?
    \item[] Answer: \answerYes{}
    \item[] Justification: We provide and check the proof for theoretical results in Section 3.

    \item {\bf Experimental result reproducibility}
    \item[] Question: Does the paper fully disclose all the information needed to reproduce the main experimental results of the paper to the extent that it affects the main claims and/or conclusions of the paper (regardless of whether the code and data are provided or not)?
    \item[] Answer: \answerNo{}
    \item[] Justification: We describe the algorithms and core settings, but the private fine-tuning data and incomplete training details prevent exact reproduction from the paper alone.

\item {\bf Open access to data and code}
    \item[] Question: Does the paper provide open access to the data and code, with sufficient instructions to faithfully reproduce the main experimental results, as described in supplemental material?
    \item[] Answer: \answerNo{}
    \item[] Justification: The fine-tuning dataset is private; the paper does not provide open access to all code and data needed for exact reproduction.

\item {\bf Experimental setting/details}
    \item[] Question: Does the paper specify all the training and test details (e.g., data splits, hyperparameters, how they were chosen, type of optimizer) necessary to understand the results?
    \item[] Answer: \answerNo{}
    \item[] Justification: We report model, data, sparsity, batch size, and training-step settings, but do not provide a complete optimizer and learning-rate specification.

\item {\bf Experiment statistical significance}
    \item[] Question: Does the paper report error bars suitably and correctly defined or other appropriate information about the statistical significance of the experiments?
    \item[] Answer: \answerNo{}
    \item[] Justification: We report point estimates without error bars, confidence intervals, or statistical significance tests.

\item {\bf Experiments compute resources}
    \item[] Question: For each experiment, does the paper provide sufficient information on the computer resources (type of compute workers, memory, time of execution) needed to reproduce the experiments?
    \item[] Answer: \answerNo{}
    \item[] Justification: We report the inference GPU and latencies, but not the full training hardware, memory, and compute budget for every experiment.

\item {\bf Code of ethics}
    \item[] Question: Does the research conducted in the paper conform, in every respect, with the NeurIPS Code of Ethics \url{https://neurips.cc/public/EthicsGuidelines}?
    \item[] Answer: \answerYes{}
    \item[] Justification: The authors have reviewed the NeurIPS Code of Ethics.

\item {\bf Broader impacts}
    \item[] Question: Does the paper discuss both potential positive societal impacts and negative societal impacts of the work performed?
    \item[] Answer: \answerNo{}
    \item[] Justification: The paper focuses on attention efficiency and does not include a dedicated discussion of societal impacts.

\item {\bf Safeguards}
    \item[] Question: Does the paper describe safeguards that have been put in place for responsible release of data or models that have a high risk for misuse (e.g., pre-trained language models, image generators, or scraped datasets)?
    \item[] Answer: \answerNA{}
    \item[] Justification: The paper does not release models, generators, or datasets with a high risk for misuse.

\item {\bf Licenses for existing assets}
    \item[] Question: Are the creators or original owners of assets (e.g., code, data, models), used in the paper, properly credited and are the license and terms of use explicitly mentioned and properly respected?
    \item[] Answer: \answerNo{}
    \item[] Justification: We cite the models and evaluation assets, but do not enumerate all asset licenses and terms of use in the paper.

\item {\bf New assets}
    \item[] Question: Are new assets introduced in the paper well documented and is the documentation provided alongside the assets?
    \item[] Answer: \answerNA{}
    \item[] Justification: The paper does not release a new dataset or model asset.

\item {\bf Crowdsourcing and research with human subjects}
    \item[] Question: For crowdsourcing experiments and research with human subjects, does the paper include the full text of instructions given to participants and screenshots, if applicable, as well as details about compensation (if any)?
    \item[] Answer: \answerNA{}
    \item[] Justification: The paper does not involve crowdsourcing nor research with human subjects.

\item {\bf Institutional review board (IRB) approvals or equivalent for research with human subjects}
    \item[] Question: Does the paper describe potential risks incurred by study participants, whether such risks were disclosed to the subjects, and whether Institutional Review Board (IRB) approvals (or an equivalent approval/review based on the requirements of your country or institution) were obtained?
    \item[] Answer: \answerNA{}
    \item[] Justification: The paper does not involve crowdsourcing nor research with human subjects.

\item {\bf Declaration of LLM usage}
    \item[] Question: Does the paper describe the usage of LLMs if it is an important, original, or non-standard component of the core methods in this research? Note that if the LLM is used only for writing, editing, or formatting purposes and does \emph{not} impact the core methodology, scientific rigor, or originality of the research, declaration is not required.
    \item[] Answer: \answerNA{}
    \item[] Justification: The core method development in this research does not involve LLMs as any important, original, or non-standard components.

\end{enumerate}

\end{document}